\documentclass[11pt]{article}
\usepackage{acl2016}
\usepackage{times}
\usepackage{latexsym}
\usepackage{caption}
\aclfinalcopy 

\title{Compression of Neural Machine Translation Models via Pruning}

\author{Abigail See$^*$ $\mbox{ }$ $\mbox{ }$ $\mbox{ }$ 
  Minh-Thang Luong\thanks{$\mbox{ }$ Both authors contributed equally.} 
 $\mbox{ }$ $\mbox{ }$ $\mbox{ }$ 
 Christopher D. Manning
 \\ Computer Science Department, Stanford University, Stanford, CA 94305 
 \\ {\tt $\{$abisee,lmthang,manning$\}$@stanford.edu }
}

\date{}

\usepackage{caption} 
\usepackage{subcaption} 
\usepackage{multirow} 
\usepackage{amsmath} 
\usepackage{amssymb} 
\usepackage{graphicx}
\usepackage{color} 
\usepackage{pstricks} 
\usepackage{arydshln} 
\usepackage{url}
\usepackage{pgfplots}

\newcommand{\hide}[1]{}

\newcommand{\tgt}[1]{y_{#1}} 
\newcommand{\src}[1]{x_{#1}} 

\newcommand{\MB}[1]{\mbox{\boldmath{$#1$}}} 
\newcommand{\open}[1]{\left(#1\right)} 

\newcommand{\bi}[1]{\textbf{\textit{#1}}}

\begin{document}

\maketitle

\begin{abstract}

Neural Machine Translation (NMT), like many other deep learning domains, typically suffers from over-parameterization, resulting in large storage sizes.
This paper examines three simple magnitude-based pruning schemes to compress NMT models, namely {\it class-blind}, {\it class-uniform}, and {\it class-distribution}, which differ in terms of how pruning thresholds are computed for the different classes of weights in the NMT architecture.
We demonstrate the efficacy of weight pruning as a compression technique for a state-of-the-art NMT system. 
We show that an NMT model with over 200 million parameters can be pruned by 40\% with very little performance loss as measured on the WMT'14 English-German translation task. 
This sheds light on the distribution of redundancy in the NMT architecture.
Our main result is that with {\it retraining}, we can recover and even surpass the original performance with an 80\%-pruned model. 


\end{abstract}

\section{Introduction}
\label{sec:intro}
\begin{figure}
\centering
\includegraphics[width=0.4\textwidth, trim = 78mm 80mm 133mm 70mm, clip]{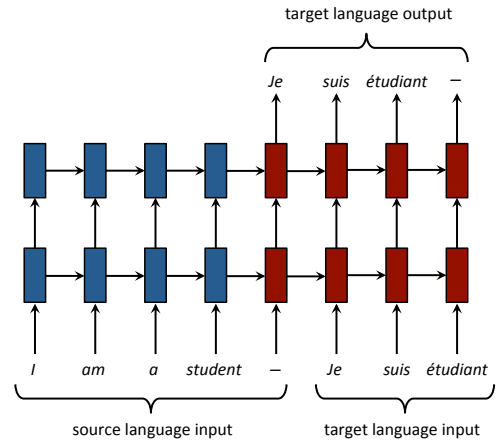}
\caption{A simplified diagram of NMT.}
\label{fig:nmt_simple}
\end{figure}

Neural Machine Translation (NMT) is a simple new architecture for translating
texts from one language into another
\cite{sutskever2014sequence,cho14}. NMT is a single deep 
neural network that is trained end-to-end, holding several advantages such as the
ability to capture long-range dependencies in
sentences, and generalization to unseen texts. Despite being relatively new, NMT has already
achieved state-of-the-art translation results for several language pairs 
including English-French \cite{luong2015addressing}, English-German
\cite{jean2015using,luong2015effective,luong15iwslt,sennrich16mono}, English-Turkish
\cite{sennrich16mono}, and English-Czech
\cite{jean15wmt,luong16char}. 
Figure~\ref{fig:nmt_simple} gives an example of an NMT system.

While NMT has a significantly smaller memory footprint than traditional phrase-based approaches (which need to store gigantic phrase-tables and
language models), the model size of NMT is still prohibitively large for mobile devices.
For example, a recent state-of-the-art NMT system requires over 200 million parameters, resulting in a storage size of hundreds of megabytes \cite{luong2015effective}. 
Though the trend for bigger and deeper neural networks has brought great progress, it has also introduced over-parameterization, resulting in long running times, overfitting, and the storage size issue discussed above. 
A solution to the over-parameterization problem could potentially aid all three issues, though the first (long running times) is outside the scope of this paper.

\bi{Our contribution.}
In this paper we investigate the efficacy of weight pruning for NMT as a means of compression.
We show that despite its simplicity, magnitude-based pruning with retraining is highly effective, and we compare three magnitude-based pruning schemes --- \textit{class-blind}, \textit{class-uniform} and \textit{class-distribution}.
Though recent work has chosen to use the latter two, we find the first and simplest scheme --- \textit{class-blind} --- the most successful.
We are able to prune 40\% of the weights of a state-of-the-art NMT system with negligible performance loss, and by adding a retraining phase after pruning, we can prune 80\% with no performance loss.
Our pruning experiments also reveal some patterns in the distribution of redundancy in NMT. In particular we find that higher layers, attention and softmax weights are the most important, while lower layers and the embedding weights hold a lot of redundancy. 
For the Long Short-Term Memory (LSTM) architecture, we find that at lower layers the parameters for the input are most crucial, but at higher layers the parameters for the gates also become important.

\section{Related Work}
\label{sec:related}
Pruning the parameters from a neural network, referred to as \textit{weight pruning} or \textit{network pruning}, is a well-established idea though it can be implemented in many ways. 
Among the most popular are the Optimal Brain Damage (OBD)
\cite{lecun1989optimal} and Optimal Brain Surgeon (OBS) \cite{hassibi1993second} techniques, which involve computing the Hessian matrix of the loss function with respect to the parameters, in order to assess the \textit{saliency} of each parameter. 
Parameters with low saliency are then pruned from the network and the remaining sparse network is retrained. 
Both OBD and OBS were shown to perform better than the so-called `naive magnitude-based approach', which prunes parameters according to their magnitude (deleting parameters close to zero).
However, the high computational complexity of OBD and OBS compare unfavorably to the computational simplicity of the magnitude-based approach, especially for large networks \cite{augasta2013pruning}.

In recent years, the deep learning renaissance has prompted a re-investigation of network pruning for modern models and tasks. 
Magnitude-based pruning with iterative retraining has yielded strong results for Convolutional Neural Networks (CNN) performing visual tasks.
\newcite{collins2014memory} prune 75\% of AlexNet parameters with small accuracy loss on the ImageNet task, while \newcite{han2015learning} prune 89\% of AlexNet parameters with no accuracy loss on the ImageNet task.

Other approaches focus on pruning neurons rather than parameters, via sparsity-inducing regularizers \cite{murray2015auto} or `wiring together' pairs of neurons with similar input weights \cite{srinivas2015data}. 
These approaches are much more constrained than weight-pruning schemes; they necessitate finding entire zero rows of weight matrices, or near-identical pairs of rows, in order to prune a single neuron. 
By contrast weight-pruning approaches allow weights to be pruned freely and independently of each other. 
The neuron-pruning approach of \newcite{srinivas2015data} was shown to perform poorly (it suffered performance loss after removing only 35\% of AlexNet parameters) compared to the weight-pruning approach of \newcite{han2015learning}. 
Though \newcite{murray2015auto} demonstrates neuron-pruning for language modeling as part of a (non-neural) Machine Translation pipeline, their approach is more geared towards architecture selection than compression.

There are many other compression techniques for neural networks, including approaches based on on low-rank approximations for weight matrices \cite{jaderberg2014speeding,denton2014exploiting}, or weight sharing  via hash functions \cite{chen2015compressing}.
Several methods involve reducing the precision of the weights or activations \cite{courbariaux2014low}, sometimes in conjunction with specialized hardware \cite{gupta2015deep}, or even using binary weights \cite{lin2015neural}.
The `knowledge distillation' technique of \newcite{hinton2015distilling} involves training a small `student' network on the soft outputs of a large `teacher' network.
Some approaches use a sophisticated pipeline of several techniques to achieve impressive feats of compression \cite{han2015deep,iandola2016squeezenet}.

\begin{figure*}[t]
\centering
\includegraphics[trim = 27mm 60mm 45mm 35mm, clip, width=\textwidth]{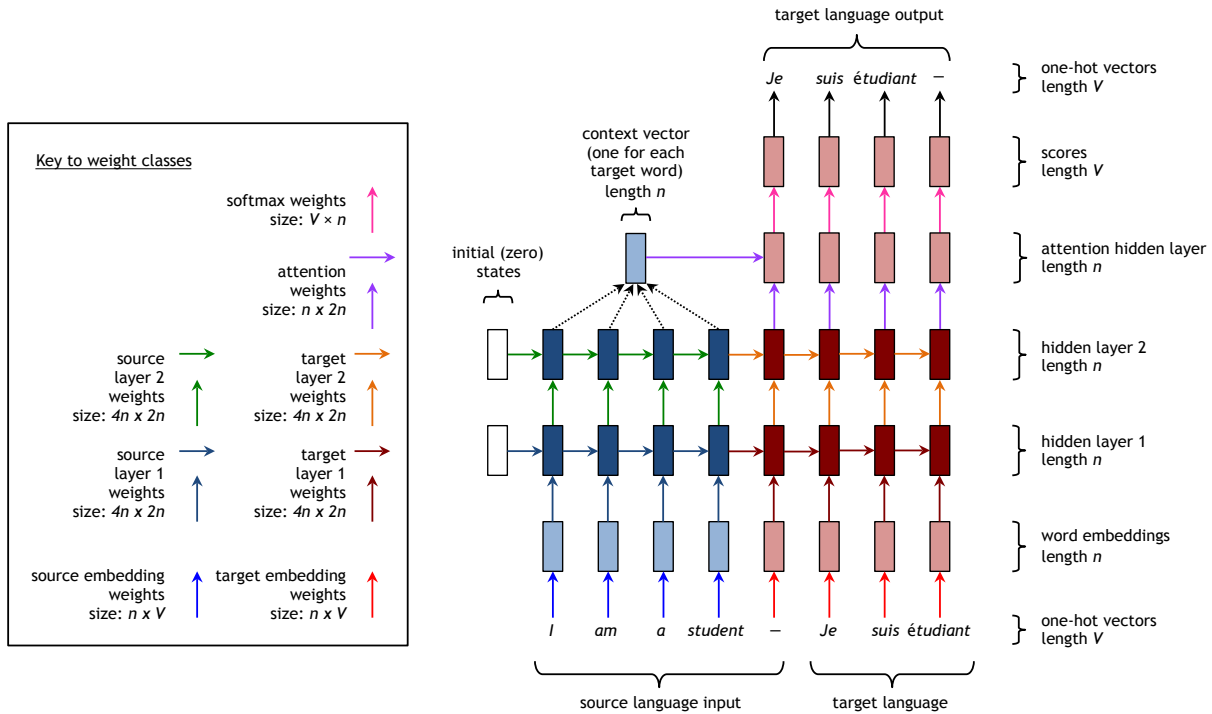}
\caption{NMT architecture. This example has two layers, but our system has four. The different weight classes are indicated by arrows of different color (the black arrows in the top right represent simply choosing the highest-scoring word, and thus require no parameters).
Best viewed in color.
}
\label{fig:nmt_complex}
\end{figure*}

Most of the above work has focused on compressing CNNs for vision tasks. 
We extend the magnitude-based pruning approach of \newcite{han2015learning} to recurrent neural networks (RNN), in particular LSTM architectures for NMT, and to our knowledge we are the first to do so.
There has been some recent work on compression for RNNs \cite{lu2016learning,prabhavalkar2016compression}, but it focuses on other, non-pruning compression techniques. 
Nonetheless, our general observations on the distribution of redundancy in a LSTM, detailed in Section \ref{subsec:redundancy}, are corroborated by \newcite{lu2016learning}.

\section{Our Approach}
\label{sec:approach}

We first give a brief overview of Neural Machine Translation before describing
the model architecture of interest, the deep multi-layer recurrent model with
LSTM. We then explain the different types of NMT weights
together with our approaches to pruning and retraining.

\subsection{Neural Machine Translation}
Neural machine translation aims to directly model the conditional probability $p(\tgt{}|\src{})$ of translating
a source sentence, $\src{1},\ldots,\src{n}$, to a target sentence, $\tgt{1},\ldots,\tgt{m}$.
It accomplishes this goal through an {\it encoder-decoder} framework
\cite{kal13,sutskever2014sequence,cho14}. The {\it encoder} computes a representation $\MB{s}$
for each source sentence. Based on that source representation,
the {\it decoder} generates a translation, one target word at a time, and hence,
decomposes the log conditional probability as:
\begin{equation}
\log p(\tgt{}|\src{}) = \sum_{t=1}^m \nolimits \log
p\open{\tgt{t}|\tgt{<t},\MB{s}}
\label{e:s2s}
\end{equation}

Most NMT work uses RNNs, but approaches differ in terms of: 
(a) architecture, which can be unidirectional, bidirectional, or deep multi-layer RNN; 
and (b) RNN type, which can be Long Short-Term Memory (LSTM) \cite{hochreiter1997long} or the Gated Recurrent Unit \cite{cho14}. 

In this work, we specifically consider the {\it deep multi-layer recurrent} architecture with {\it
LSTM} as the hidden unit type.
Figure \ref{fig:nmt_simple} illustrates an instance of that architecture during training in which the source and target sentence pair are input for supervised
learning. During testing, the target sentence is not known in advance; instead, the most probable
target words predicted by the model are fed as inputs into the next timestep.
The network stops when it emits the end-of-sentence symbol --- a special `word' in the vocabulary, represented by a dash in Figure \ref{fig:nmt_simple}.

\subsection{Understanding NMT Weights}
\label{subsec:lstm}
Figure~\ref{fig:nmt_complex} shows the same system in more detail,
highlighting the different types of parameters, or weights, in the model.
We will go through the architecture from bottom to top.
First, a vocabulary is chosen for each language, assuming that the top $V$ frequent
words are selected.
Thus, every word in the source or target vocabulary can be represented by a one-hot vector of length $V$.
The source input sentence and target input sentence, represented as a sequence
of one-hot vectors, are transformed into a sequence of word embeddings by the
\emph{embedding} weights. 
These embedding weights, which are learned during training, are different for the source words and the target words.
The word embeddings and all hidden layers are vectors of length $n$ (a chosen hyperparameter).

The word embeddings are then fed as input into the main network, which consists
of two multi-layer RNNs `stuck together' --- an encoder for the source
language and a decoder for the target language, each with their own
weights. 
The \emph{feed-forward} (vertical) weights connect
the hidden unit from the layer below to the upper RNN block, and the
\emph{recurrent} (horizontal) weights connect the hidden unit from the previous
time-step RNN block to the current time-step RNN block.

The hidden state at the top layer of the decoder is fed through an
\textit{attention} layer, which guides the translation by `paying attention' to relevant parts of the source sentence; 
for more information see \newcite{bahdanau2014neural} or Section 3 of \newcite{luong2015effective}.
Finally, for each target word, the top layer hidden unit is transformed by the
\emph{softmax} weights into a score vector of length $V$. The target word with the highest score is selected as the output translation.

\bi{Weight Subgroups in LSTM} -- For the aforementioned RNN block, we choose to
use LSTM as the hidden unit type. To facilitate our later discussion 
on the different subgroups of weights
within LSTM, we first review the details of LSTM as formulated by 
\newcite{zaremba2014recurrent} as follows:
\begin{align}
\begin{pmatrix}
i\\
f\\
o\\
\hat{h}
\end{pmatrix}
&=
\begin{pmatrix}
\text{sigm}\\
\text{sigm}\\
\text{sigm}\\
\text{tanh}
\end{pmatrix}
T_{4n,2n}
\begin{pmatrix}
h_t^{l-1}\\
h_{t-1}^l
\end{pmatrix} \label{eqn:lstm_1} \\
c_t^l&=f \circ c_{t-1}^l + i \circ \hat{h} \label{eqn:lstm_2} \\
h_t^l &= o \circ \text{tanh}(c_t^l) \label{eqn:lstm_3}
\end{align}
Here, each LSTM block at time $t$ and layer $l$ computes as output a pair of
hidden and memory vectors ($h_t^l$, $c_t^l$) given the previous pair
($h_{t-1}^l$, $c_{t-1}^l$) and an input vector $h_t^{l-1}$ (either from the LSTM block below or
the embedding weights if $l\!=\!1$). All of these vectors
have length $n$.

The core of a LSTM block is the weight matrix $T_{4n,2n}$ of size $4n \times
2n$. This matrix can be decomposed into 8 subgroups that are responsible for the
interactions between $\{$input gate $i$, forget gate $f$, output gate $o$,
input signal $\hat{h}\} \times \{$feed-forward input $h_t^{l-1}$, recurrent
input $h_{t-1}^l\}$.

%

\subsection{Pruning Schemes}
\label{subsec:approach_schemes}
We follow the general magnitude-based approach of \newcite{han2015learning}, which consists of pruning weights with smallest absolute value. However, we question the authors' pruning scheme with respect to the different weight classes, and experiment with three pruning schemes.
Suppose we wish to prune $x$\% of the total parameters in the model. 
How do we distribute the pruning over the different weight classes (illustrated in Figure~\ref{fig:nmt_complex}) of our model? 
We propose to examine three different pruning schemes:
\begin{enumerate}
\item \textit{Class-blind}: 
Take all parameters, sort them by magnitude and prune the $x$\% with smallest magnitude, regardless of weight class.
(So some classes are pruned proportionally more than others).
\item \textit{Class-uniform}: 
Within each class, sort the weights by magnitude and prune the $x$\% with smallest magnitude.
(So all classes have exactly $x$\% of their parameters pruned).
\item \textit{Class-distribution}: 
 For each class $c$, weights with magnitude less than $\lambda \sigma_c$ are
 pruned. Here, $\sigma_c$ is the standard deviation of that class and $\lambda$ is a universal parameter chosen such that in total, $x\%$ of all parameters are pruned.
This is used by \newcite{han2015learning}.
\end{enumerate}
All these schemes have their seeming advantages.
Class-blind pruning is the simplest and adheres to the principle that pruning
weights (or equivalently, setting them to zero) is least damaging when
those weights are small, regardless of their locations in the architecture.
Class-uniform pruning and class-distribution pruning both seek to prune
proportionally within each weight class, either absolutely, or relative to the
standard deviation of that class.
We find that class-blind pruning outperforms both other schemes (see Section~\ref{subsec:exp_schemes}).

\subsection{Retraining}
\label{subsec:approach_retraining}
In order to prune NMT models aggressively without performance loss, we retrain our pruned networks. 
That is, we continue to train the remaining weights, but maintain the sparse structure introduced by pruning.
In our implementation, pruned weights are represented by zeros in the weight matrices, 
and we use binary `mask' matrices, which represent the sparse structure of a network, 
to ignore updates to weights at pruned locations.
This implementation has the advantage of simplicity as it requires minimal changes to the training and deployment code, 
but we note that a more complex implementation utilizing sparse matrices and sparse matrix multiplication could potentially yield speed improvements.
However, such an implementation is beyond the scope of this paper.

\begin{figure}
\centering

\begin{tikzpicture}

\begin{axis}[%
width=0.8\columnwidth,
height=4cm,
scale only axis,
xmin=0,
xmax=90,
xtick={0, 10, 20, 30, 40, 50, 60, 70, 80, 90},
xlabel={percentage pruned},
ymin=0,
ymax=25,
yminorticks=true,
ylabel={BLEU score},
axis background/.style={fill=white},
legend pos = south west,
legend cell align=left
]
\addplot [color=blue,solid,mark=x,mark options={solid}]
  table[row sep=crcr]{%
  0	20.48\\
10	20.44\\
20	20.44\\
30	20.31\\
40	20.15\\
50	19.55\\
60	18.81\\
70	16.41\\
80	10.99\\
90	1.2\\
};
\addlegendentry{class-blind}

\addplot [color=red,solid,mark=x,mark options={solid}]
  table[row sep=crcr]{%
    0	20.48\\
10	20.45\\
20	20.17\\
30	20.19\\
40	19.25\\
50	18.05\\
60	14.99\\
70	9.64\\
80	3.18\\
90	0.35\\
};
\addlegendentry{class-uniform}

\addplot [color=green,solid,mark=x,mark options={solid}]
  table[row sep=crcr]{%
    0	20.48\\
10	20.43\\
20	20.19\\
30	19.95\\
40	19.41\\
50	17.8\\
60	15.02\\
70	9.71\\
80	3.03\\
90	0.41\\
};
\addlegendentry{class-distribution}

\addplot [color=black,dashed]
  table[row sep=crcr]{%
0	20.48\\
90	20.48\\
};
\end{axis}
\end{tikzpicture}%
\caption{Effects of different pruning schemes.}
\label{fig:pruning_methods}
\end{figure}
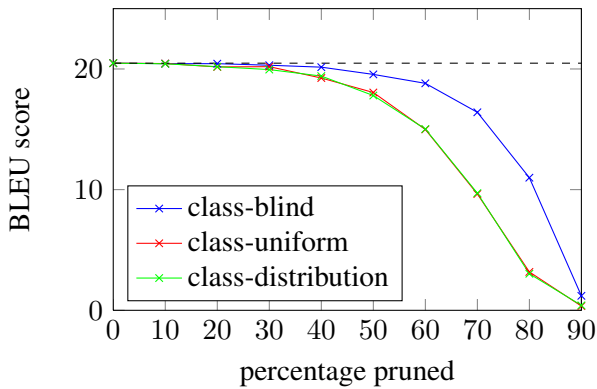

\begin{figure*}
\centering
\begin{tikzpicture}
    \begin{axis}[
width=\textwidth,
height=6cm,
        major x tick style = transparent,
        ybar,
        bar width=4.5pt,
        ymajorgrids = true,
        ylabel = {perplexity change},
        symbolic x coords={source layer 1, source layer 2, source layer 3, source layer 4, target layer 1, target layer 2, target layer 3, target layer 4, attention, softmax, source embedding, target embedding},
        x tick label style={rotate=45, anchor=north east, inner sep=0mm}, 
        xtick = data,
        x=30pt, 
        scaled y ticks = false,
     legend pos = north west,   
     legend cell align = left,
    ]
        \addplot[style={blue!50,fill=blue!50,mark=none}]
            coordinates {(source layer 1,0.46659)  (source layer 2,0.362434)  (source layer 3,0.796254)  (source layer 4,0.794582)  (target layer 1,0.201001)  (target layer 2,0.222658)  (target layer 3,-0.291916)  (target layer 4,1.108432)  (attention,0.4164)  (softmax,7.803921)  (source embedding,2.962925)  (target embedding,2.351362)};
            
        \addplot[style={red!50,fill=red!50,mark=none}]
            coordinates {(source layer 1,0.471137)  (source layer 2,0.561763)  (source layer 3,1.829927)  (source layer 4,3.628541)  (target layer 1,0.422055)  (target layer 2,0.457759)  (target layer 3,0.473893)  (target layer 4,7.545939)  (attention,6.362093)  (softmax,16.277336)  (source embedding,0.335614)  (target embedding,0.226382) };

        \addplot[style={green!50,fill=green!50,mark=none}]
            coordinates {(source layer 1,0.458798)  (source layer 2,0.543843)  (source layer 3,1.818222)  (source layer 4,3.93477)  (target layer 1,0.428267)  (target layer 2,0.46657)  (target layer 3,0.494273)  (target layer 4,8.308794)  (attention,6.34294)  (softmax,15.723167)  (source embedding,0.311599)  (target embedding,0.253745) 
 };
        \legend{class-blind, class-uniform, class-distribution}
    \end{axis}
\end{tikzpicture}
\caption{`Breakdown' of performance loss (i.e., perplexity increase) by weight class, when pruning 90\% of weights using each of the three pruning schemes. Each of the first eight classes have 8 million weights, attention has 2 million, and the last three have 50 million weights each.}
\label{fig:breakdown}
\end{figure*}
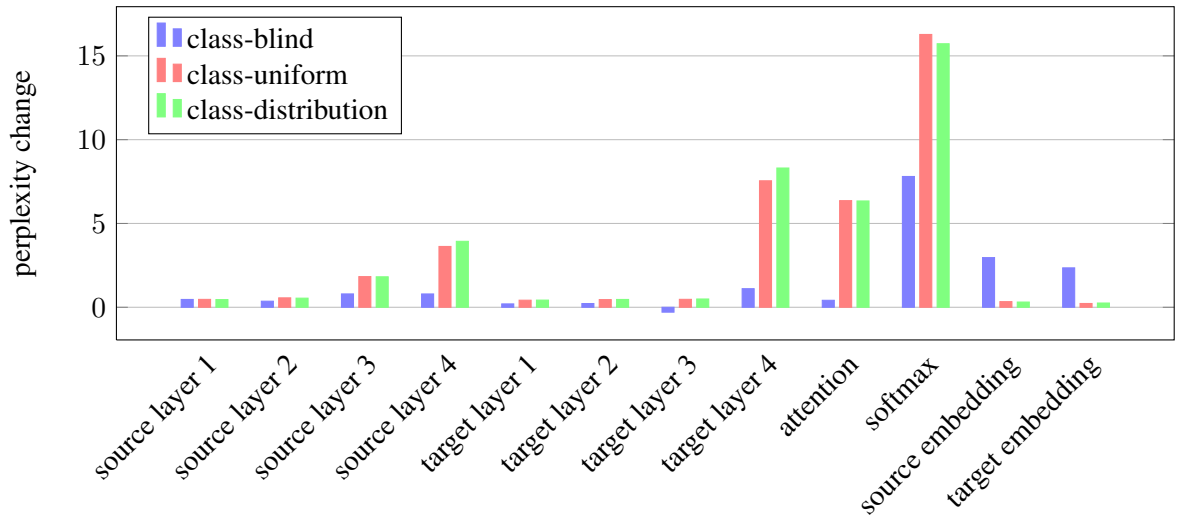
\section{Experiments}
\label{sec:exp}

We evaluate the effectiveness of our pruning approaches on a state-of-the-art
NMT model.\footnote{We thank the authors of \newcite{luong2015effective}
for providing their trained models and assistance in using the codebase
at \url{https://github.com/lmthang/nmt.matlab}.} 
Specifically, an attention-based English-German NMT system from
\newcite{luong2015effective} is considered. 
Training data was obtained from WMT'14 consisting
of 4.5M sentence pairs (116M English words, 110M German words). For
more details on training hyperparameters, we refer readers to Section 4.1 of
\newcite{luong2015effective}.
All models are tested on newstest2014 (2737 sentences). 
The model achieves a
perplexity of 6.1 and a BLEU score of
20.5 (after unknown word replacement).\footnote{The performance of this model
is reported under row {\it global (dot)} in Table 4 of
\newcite{luong2015effective}.}

When {\it retraining} pruned NMT systems, we use the following settings: (a) we start
with a smaller learning rate of 0.5 (the original model uses a learning rate of
1.0), (b) we train for fewer epochs, 4 instead of 12, using plain SGD, (c) a simple learning
rate schedule is employed; after 2 epochs, we begin to halve the learning rate
every half an epoch, and (d) all other hyperparameters are the same, such as
mini-batch size 128, maximum gradient norm 5, and dropout with probability 0.2.

\subsection{Comparing pruning schemes}
\label{subsec:exp_schemes}
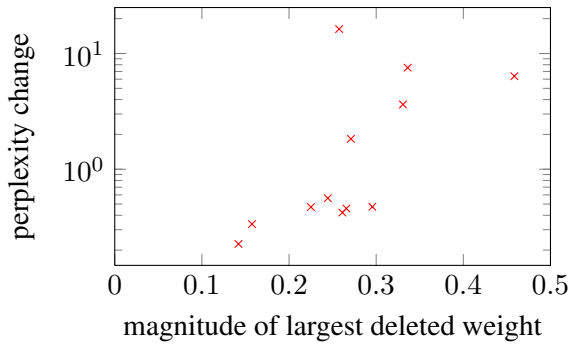
\begin{figure}
\centering

\begin{tikzpicture}
\begin{semilogyaxis}[
width=0.95\columnwidth,
height=5cm,
xlabel={magnitude of largest deleted weight}, 
ylabel={perplexity change},
xmin=0,
xmax=0.5,
]
\addplot[
only marks,
color=red,
mark=x,
]
table[row sep=crcr]
{
0.22513		0.471137\\
0.244492		0.561763\\
0.270999		1.829927\\
0.330704		3.628541\\
0.261247		0.422055\\
0.265832		0.457759\\
0.295644		0.473893\\
0.336088		7.545939\\
0.458583		6.362093\\
0.257479		16.277336\\
0.157486		0.335614\\
0.14194		0.226382\\
};
\end{semilogyaxis}
\end{tikzpicture}
\caption{Magnitude of largest deleted weight vs. perplexity change, for the 12 different weight classes when pruning 90\% of parameters by class-uniform pruning.}
\label{fig:scatter}
\end{figure}

Despite its simplicity, we observe in Figure~\ref{fig:pruning_methods} that {\it
class-blind} pruning outperforms both other schemes in terms of translation
quality at all pruning percentages.
In order to understand this result, for each of the three pruning schemes, we pruned each class separately and recorded the effect on performance (as measured by perplexity).
Figure \ref{fig:breakdown} shows that with class-uniform pruning, the overall performance loss is caused disproportionately by a few classes: target layer 4, attention and softmax weights. Looking at Figure \ref{fig:scatter}, we see that the most damaging classes to prune also tend to be those with weights of greater magnitude --- these classes have much larger weights than others at the same percentile, so pruning them under the class-uniform pruning scheme is more damaging. The situation is similar for class-distribution pruning.

By contrast, Figure \ref{fig:breakdown} shows that under class-blind pruning, the damage caused by pruning softmax, attention and target layer 4 weights is greatly decreased, and the contribution of each class towards the performance loss is overall more uniform.
In fact, the distribution begins to reflect the number of parameters in each class --- for example, the source and target embedding classes have larger contributions because they have more weights. 
We use only class-blind pruning for the rest of the experiments.

Figure \ref{fig:breakdown} also reveals some interesting information about the distribution of redundancy in NMT architectures --- namely it seems that higher layers are more important than lower layers, and that attention and softmax weights are crucial. We will explore the distribution of redundancy further in Section \ref{subsec:redundancy}.

\subsection{Pruning and retraining}
\label{sec:effect}

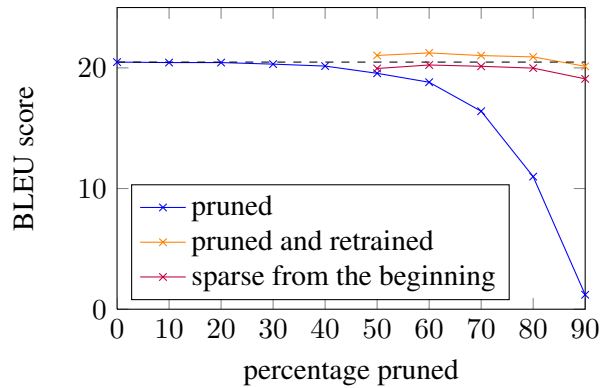
\begin{figure}
\centering

\begin{tikzpicture}

\begin{axis}[%
width=0.8\columnwidth,
height=4cm,
scale only axis,
xmin=0,
xmax=90,
xtick={0,10, 20, 30, 40, 50, 60, 70, 80, 90},
xlabel={percentage pruned},
ymin=0,
ymax=25,
yminorticks=true,
ylabel={BLEU score},
axis background/.style={fill=white},
legend pos = south west,
legend cell align=left,
]
\addplot [color=blue,solid,mark=x,mark options={solid}]
  table[row sep=crcr]{%
  0	20.48\\
10	20.44\\
20	20.44\\
30	20.31\\
40	20.15\\
50	19.55\\
60	18.81\\
70	16.41\\
80	10.99\\
90	1.2\\
};
\addlegendentry{pruned}

\addplot [color=orange,solid,mark=x,mark options={solid}]
  table[row sep=crcr]{%
50	21.03\\
60	21.24\\
70	21.02\\
80	20.91\\
90	20.13\\
};
\addlegendentry{pruned and retrained}

\addplot [color=purple,solid,mark=x,mark options={solid}]
  table[row sep=crcr]{%
50	19.95\\
60	20.24\\
70	20.13\\
80	19.98\\
90	19.09\\
};
\addlegendentry{sparse from the beginning}

\addplot [color=black,dashed]
  table[row sep=crcr]{%
0	20.48\\
90	20.48\\
};
\end{axis}
\end{tikzpicture}%
\caption{Performance of pruned models (a) after pruning, (b) after pruning and retraining, and (c) when trained with sparsity structure from the outset (see Section \ref{sec:sparse}).}
\label{fig:main_results}
\end{figure}

Pruning has an immediate negative impact on performance (as measured by BLEU) that is exponential in pruning percentage; this is demonstrated by the blue line in Figure \ref{fig:main_results}.
However we find that up to about 40\% pruning, performance is mostly unaffected, indicating a large amount of redundancy and over-parameterization in NMT.

We now consider the effect of retraining pruned models.
The orange line in Figure \ref{fig:main_results} shows that after retraining the pruned models, baseline performance (20.48 BLEU) is both recovered and improved upon, up to 80\% pruning (20.91 BLEU), with only a small performance loss at 90\% pruning (20.13 BLEU).
This may seem surprising, as we might not expect a sparse model to significantly out-perform a model with five times as many parameters.
There are several possible explanations, two of which are given below.

Firstly, we found that the less-pruned models perform better on the training set than the validation set, whereas the more-pruned models have closer performance on the two sets. 
This indicates that pruning has a regularizing effect on the retraining phase, though clearly more is not always better, as the 50\% pruned and retrained model has better validation set performance than the 90\% pruned and retrained model.
Nonetheless, this regularization effect may explain why the pruned and retrained models outperform the baseline.
\begin{figure}[tbh]
\centering

\begin{tikzpicture}

\begin{axis}[%
width=0.8\columnwidth,
height=4cm,
scale only axis,
xmin=0,
xmax=540000,
xlabel={training iterations},
ymin=1,
ymax=8,
yminorticks=true,
ylabel={loss},
axis background/.style={fill=white},
]

\addplot [color=black,dotted]
  table[row sep=crcr]{%
0	2.27\\
540000	2.27\\
};

\addplot [color=black,dotted]
  table[row sep=crcr]{%
405000	0\\
405000	8\\
};

\addplot [color=blue,solid,mark options={solid}]
  table[row sep=crcr]{%
5000	6.89\\
10000	5.91\\
15000	5.34\\
20000	4.88\\
25000	4.35\\
30000	3.9\\
35000	3.66\\
40000	3.61\\
45000	3.55\\
50000	3.5\\
55000	3.43\\
60000	3.3\\
65000	3.16\\
70000	3.07\\
75000	3.06\\
80000	3.05\\
85000	3.04\\
90000	3.01\\
95000	2.94\\
100000	2.86\\
105000	2.82\\
110000	2.82\\
115000	2.82\\
120000	2.81\\
125000	2.8\\
130000	2.75\\
135000	2.7\\
140000	2.67\\
145000	2.67\\
150000	2.67\\
155000	2.67\\
160000	2.66\\
165000	2.63\\
170000	2.59\\
175000	2.57\\
180000	2.57\\
185000	2.57\\
190000	2.57\\
195000	2.57\\
200000	2.54\\
205000	2.51\\
210000	2.5\\
215000	2.5\\
220000	2.5\\
225000	2.5\\
230000	2.5\\
235000	2.48\\
240000	2.45\\
245000	2.44\\
250000	2.44\\
255000	2.44\\
260000	2.44\\
265000	2.44\\
270000	2.42\\
275000	2.4\\
280000	2.39\\
285000	2.39\\
290000	2.4\\
295000	2.4\\
300000	2.4\\
305000	2.38\\
310000	2.36\\
315000	2.35\\
320000	2.35\\
325000	2.36\\
330000	2.36\\
335000	2.35\\
340000	2.34\\
345000	2.32\\
350000	2.31\\
355000	2.31\\
360000	2.31\\
365000	2.31\\
370000	2.31\\
375000	2.29\\
380000	2.28\\
385000	2.27\\
390000	2.27\\
395000	2.27\\
400000	2.27\\
405000	2.27\\
410000	2.53\\
415000	2.52\\
420000	2.48\\
425000	2.42\\
430000	2.22\\
435000	2.05\\
440000	1.99\\
445000	2.04\\
450000	2.08\\
455000	2.11\\
460000	2.12\\
465000	2.06\\
470000	1.99\\
475000	1.96\\
480000	1.99\\
485000	2.02\\
490000	2.03\\
495000	2.04\\
500000	2.01\\
505000	1.96\\
510000	1.95\\
515000	1.97\\
520000	1.98\\
525000	2\\
530000	2\\
535000	1.98\\
540000	1.95\\
};

\end{axis}
\end{tikzpicture}%
\caption{The validation set loss during training, pruning and retraining. The vertical dotted line marks the point when 80\% of the parameters are pruned. The horizontal dotted line marks the best performance of the unpruned baseline.}
\label{fig:loss_curve}
\end{figure}

\begin{figure*}
\centering
\includegraphics[trim = 0mm 130mm 300mm 0mm, clip, width=\textwidth]{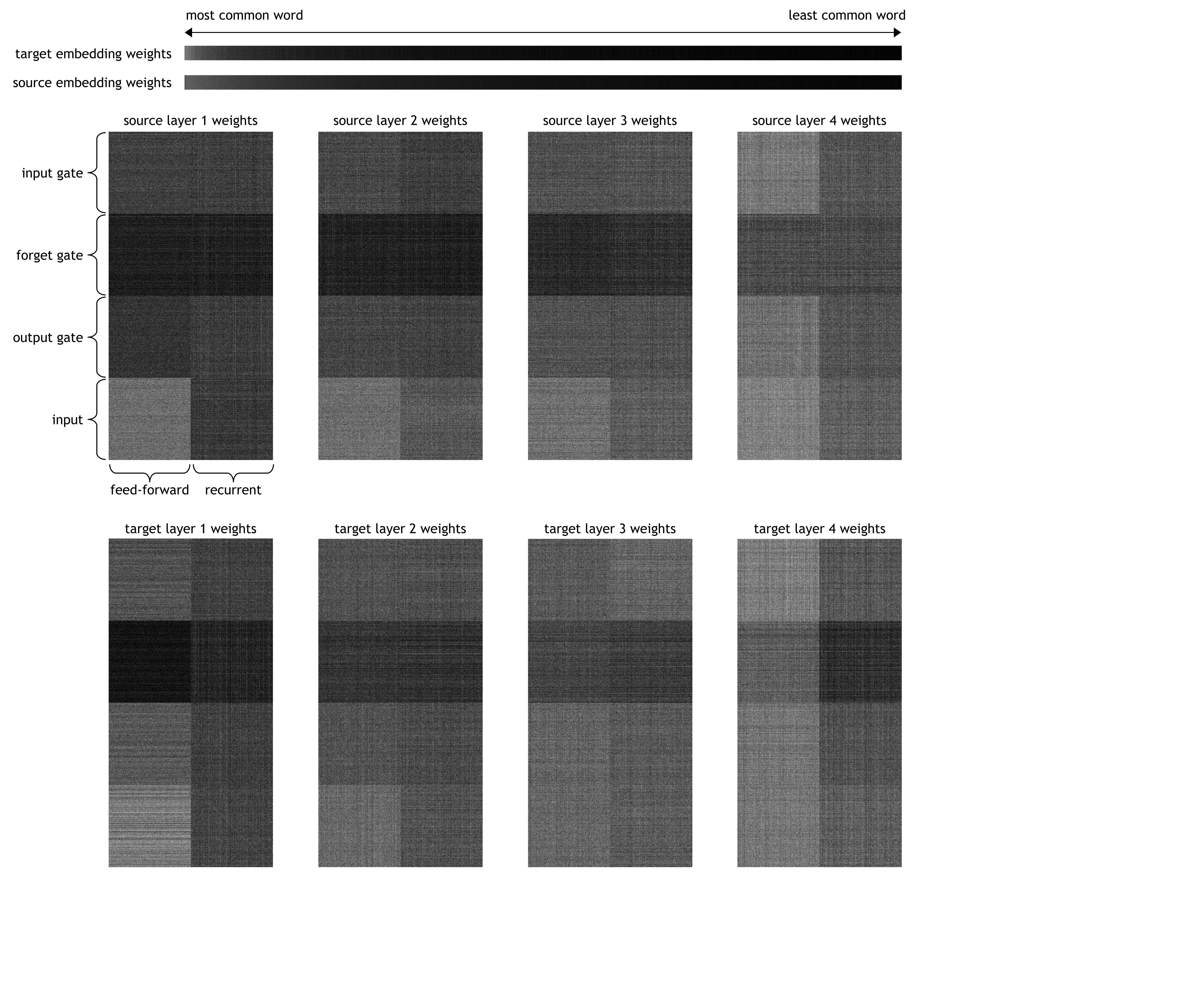}
\caption{Graphical representation of the location of small weights in various parts of the model. 
Black pixels represent weights with absolute size in the bottom 80\%; white pixels represent those with absolute size in the top 20\%.
Equivalently, these pictures illustrate which parameters remain after pruning 80\% using our class-blind pruning scheme.
}
\label{fig:redundancy_location}
\end{figure*}

Alternatively, pruning may serve as a means to escape a local optimum. 
Figure \ref{fig:loss_curve} shows the loss function over time during the training, pruning and retraining process.
During the original training process, the loss curve flattens out and seems to converge (note that we use early stopping to obtain our baseline model, so the original model was trained for longer than shown in Figure \ref{fig:loss_curve}).
Pruning causes an immediate increase in the loss function, but enables further gradient descent, allowing the retraining process to find a new, better local optimum.
It seems that the disruption caused by pruning is beneficial in the long-run.

\subsection{Starting with sparse models}
\label{sec:sparse}
The favorable performance of the pruned and retrained models raises the question: can we get a shortcut to this performance by \emph{starting} with sparse models?
That is, rather than train, prune, and retrain, what if we simply prune then train?
To test this, we took the sparsity structure of our 50\%--90\% pruned models, and trained completely new models with the same sparsity structure.
The purple line in Figure \ref{fig:main_results} shows that the `sparse from the beginning' models do not perform as well as the pruned and retrained models, but they do come close to the baseline performance.
This shows that while the sparsity structure alone contains useful information about redundancy and can therefore produce a competitive compressed model, it is important to interleave pruning with training.

Though our method involves just one pruning stage, other pruning methods interleave pruning with training more closely by including several iterations \cite{collins2014memory,han2015learning}.
We expect that implementing this for NMT would likely result in further compression and performance improvements.

\subsection{Storage size}
The original unpruned model (a MATLAB file) has size 782MB.
The 80\% pruned and retrained model is 272MB, which is a 65.2\% reduction.
In this work we focus on compression in terms of number of parameters rather than storage size, because it is invariant across implementations.

\subsection{Distribution of redundancy in NMT}
\label{subsec:redundancy}

We visualize in Figure~\ref{fig:redundancy_location} the redundancy structore of
our NMT baseline model.
{\it Black} pixels represent weights near to zero (those that can be pruned); {\it white} pixels represent larger ones.
First we consider the embedding weight matrices, whose columns correspond to words in the vocabulary.
Unsurprisingly, in Figure \ref{fig:redundancy_location}, we see that the parameters corresponding to the less common words are more dispensable.
In fact, at the 80\% pruning rate, for 100 uncommon source words and 1194
uncommon target words, we delete \emph{all} parameters corresponding to that word.
This is not quite the same as removing the word from the vocabulary --- true out-of-vocabulary words are mapped to the embedding for the `unknown word' symbol, whereas these `pruned-out' words are mapped to a zero embedding.
However in the original unpruned model these uncommon words already had near-zero embeddings, indicating that the model was unable to learn sufficiently distinctive representations.

Returning to Figure \ref{fig:redundancy_location}, now look at the eight weight matrices for the source and target connections at each of the four layers.
Each matrix corresponds to the $4n \times 2n$ matrix $T_{4n,2n}$ in Equation (\ref{eqn:lstm_1}).
In all eight matrices, we observe --- as does \newcite{lu2016learning} --- that the weights connecting to the input $\hat{h}$ are most crucial, followed by the input gate $i$, then the output gate $o$, then the forget gate $f$. 
This is particularly true of the lower layers, which focus primarily on the input $\hat{h}$. 
However for higher layers, especially on the target side, weights connecting to the gates are as important as those connecting to the input $\hat{h}$.
The gates represent the LSTM's ability to add to, delete from or retrieve information from the memory cell.
Figure \ref{fig:redundancy_location} therefore shows that these sophisticated memory cell abilities are most important at the \emph{end} of the NMT pipeline (the top layer of the decoder).
This is reasonable, as we expect higher-level features to be learned later in a deep learning pipeline.

We also observe that for lower layers, the feed-forward input is much more important than the recurrent input, whereas for higher layers the recurrent input becomes more important.
This makes sense: lower layers concentrate on the low-level information from the current word embedding (the feed-forward input), whereas higher layers make use of the higher-level representation of the sentence so far (the recurrent input).

Lastly, on close inspection, we notice several white diagonals emerging within
some subsquares of the matrices in Figure \ref{fig:redundancy_location},
indicating that even without initializing the weights to identity matrices
(as is sometimes done \cite{le2015simple}),
an identity-like weight matrix is learned. At higher pruning percentages, these diagonals become more pronounced.

\section{Generalizability of our results}
To test the generalizability of our results, we also test our pruning approach
on a smaller, non-state-of-the-art NMT model trained on the WIT3 Vietnamese-English 
dataset \cite{cettoloEtAl:EAMT2012}, which consists of 133,000 sentence pairs.
This model is effectively a scaled-down version of the state-of-the-art model in \newcite{luong2015effective},
with fewer layers, smaller vocabulary size, smaller hidden layer size, no attention mechanism,
and about 11\% as many parameters in total.
It achieves a BLEU score of 9.61 on the validation set.

Although this model and its training set are on a different scale to our main model, 
and the language pair is different, 
we found very similar results. 
For this model, it is possible to prune 60\% of parameters with no immediate performance loss,
and with retraining it is possible to prune 90\%, and regain original performance.
Our main observations from Sections \ref{subsec:exp_schemes} to \ref{subsec:redundancy}
are also replicated; in particular, class-blind pruning is most successful,
`sparse from the beginning' models are less successful than pruned and retrained models,
and we observe the same patterns as seen in Figure \ref{fig:redundancy_location}.

\section{Future Work}
As noted in Section \ref{sec:sparse}, including \emph{several} iterations of pruning and retraining would likely improve the compression and performance of our pruning method.
If possible it would be highly valuable to exploit the sparsity of the pruned models to speed up training and runtime, perhaps through sparse matrix representations and multiplications (see Section \ref{subsec:approach_retraining}).
Though we have found magnitude-based pruning to perform very well, it would be instructive to revisit the original claim that other pruning methods (for example Optimal Brain Damage and Optimal Brain Surgery) are more principled, and perform a comparative study.

\section{Conclusion}
\label{sec:conclusion}
We have shown that weight pruning with retraining is a highly effective method of compression and regularization on a state-of-the-art NMT system, compressing the model to 20\% of its size with no loss of performance. 
Though we are the first to apply compression techniques to NMT, we obtain a similar degree of compression to other current work on compressing state-of-the-art deep neural networks, with an approach that is simpler than most.
We have found that the absolute size of parameters is of primary importance when choosing which to prune, leading to an approach that is extremely simple to implement, and can be applied to any neural network.
Lastly, we have gained insight into the distribution of redundancy in the NMT architecture.

\section{Acknowledgment}
This work was partially supported by NSF Award IIS-1514268 and partially supported by a gift from Bloomberg L.P.
We gratefully acknowledge the support of the Defense Advanced Research Projects Agency (DARPA) Communicating with Computers (CwC) program under ARO prime contract no. W911NF-15-1-0462.
Lastly, we acknowledge NVIDIA Corporation for the donation of Tesla K40 GPUs.

\newpage
\bibliography{acl2016}
\bibliographystyle{acl2016}

\end{document}